%% file: zeller2024ral.tex
\documentclass[letterpaper, 10 pt, journal, twoside]{IEEEtran}

\input{stachnisslab-latex}

\input{stachnisslab-math}

\usepackage{hyperref} 

\usepackage{multirow}
\usepackage{multicol}
\usepackage{graphicx}
\usepackage{siunitx}
\usepackage{booktabs}
\usepackage[table,xcdraw]{xcolor}
\title{SemRaFiner: Panoptic Segmentation \\ in Sparse and Noisy Radar Point Clouds}

\author{ \quad \,\, Matthias Zeller$^{1}$ \qquad \qquad \,  \quad\qquad  \and Daniel Casado Herraez$^{2}$ \qquad \quad \qquad  \quad \qquad \and Bengisu Ayan$^{3}$ \qquad\quad \, \,\quad  \negthinspace \and  Jens Behley$^{4}$ \qquad   \qquad  \enspace \quad \quad \enspace \, \and Michael Heidingsfeld$^{5}$  \qquad \quad \quad\quad\qquad   \negthinspace \and Cyrill Stachniss$^{6}$%
  \thanks{Manuscript received: June 6, 2024; Revised: Aug. 28, 2022; Accepted: Oct. 7, 2024. 
  This paper was recommended for publication by Editor Hyungpil Moon upon evaluation of the Associate Editor and Reviewers' comments.}%
  
  \thanks{$^{1}$Matthias Zeller and $^{2}$Daniel Casado Herraez are with CARIAD SE and with the Center for Robotics, University of Bonn, Germany. $^{3}$Bengisu Ayan is with CARIAD SE and with the Technical University of Munich, Germany. $^{4}$Jens Behley is with the Center for Robotics, University of Bonn, Germany. $^{5}$Michael Heidingsfeld is with CARIAD SE, Germany. $^{6}$Cyrill Stachniss is with the Center for Robotics, University of Bonn, Germany, and with the Lamarr Institute for Machine Learning and Artificial Intelligence, Germany. Corresponding author: {\tt\footnotesize matthias.zeller@cariad.technology}}%
  \thanks{Digital Object Identifier (DOI): see top of this page.}
}

\begin{document}
\maketitle
\markboth{IEEE Robotics and Automation Letters. Preprint Version. Accepted October, 2024.}
{Zeller \MakeLowercase{\textit{et al.}}: SemRaFiner: Panoptic Segmentation in Sparse and Noisy Radar Point Clouds}

\begin{abstract}

  Semantic scene understanding, including the perception and classification of moving agents, is essential to enabling safe and robust driving behaviours of autonomous vehicles. Cameras and LiDARs are commonly used for semantic scene understanding. However, both sensor modalities face limitations in adverse weather and usually do not provide motion information. Radar sensors overcome these limitations and directly offer information about moving agents by measuring the Doppler velocity, but the measurements are comparably sparse and noisy.
  In this paper, we address the problem of panoptic segmentation in sparse radar point clouds to enhance scene understanding. Our approach, called SemRaFiner, accounts for changing density in sparse radar point clouds and optimizes the feature extraction to improve accuracy. Furthermore, we propose an optimized training procedure to refine instance assignments by incorporating a dedicated data augmentation.  
  Our experiments suggest that our approach outperforms state-of-the-art methods for radar-based panoptic segmentation.
\end{abstract}

\begin{IEEEkeywords}
Semantic Scene Understanding, Deep Learning Methods
\end{IEEEkeywords}
\section{Introduction}
\label{sec:intro}

\IEEEPARstart{A}{utonomous} vehicles require reliable perception and semantic scene understanding to enable safe driving behaviour. In particular, dynamic traffic participants must be correctly identified to ensure secure autonomous mobility. While cameras and LiDARs enhance the overall scene understanding, these sensors face limitations under adverse weather. Radar sensors work under adverse weather conditions, including rain, fog and snow, overcoming the shortcomings of cameras and LiDARs. Furthermore, radar sensors provide additional information, such as the Doppler velocity, the relative radial velocity between the sensor and the detected object, and the radar cross section~(RCS) value, which depends on the surface, the material and the shape of the target. However, radar measurements are sparse and affected by noise due to ego-motion and multi-path propagation.

\begin{figure}[t]
  \centering
  \fontsize{10pt}{10pt}\selectfont
     \def\svgwidth{\linewidth}
     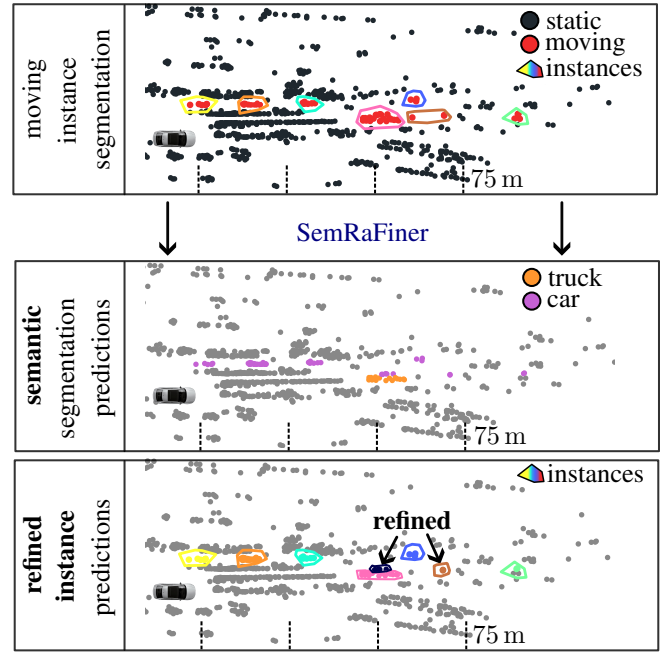
  \caption{Our method processes moving instance segmentation predictions (a) to predict point-wise semantic classes (b) and refine the instance assignment (c) to solve the panoptic segmentation task from sparse radar point clouds. In the image of the point cloud (a), each polygon represents a different instance of moving objects.}
  \label{fig:motivation}
  \vspace{-0.4cm}
\end{figure}
In this paper, we tackle the problem of panoptic segmentation, combining instance and semantic segmentation, in sparse and noisy radar point clouds. We focus on moving instances, which first require the differentiation of the moving and static parts of the environment, such as parked and moving vehicles. In the second step, we separate the moving agents into individual instances within the surrounding and assign them a corresponding semantic class. This incorporates refining the semantic prediction to enhance the scene understanding, resulting in panoptic segmentation.
Radar sensors are particularly suitable for identifying moving agents because they provide the Doppler velocity of the detection. We leverage moving instance predictions and propose a semantic instance refinement illustrated in~\figref{fig:motivation}. Compared to state-of-the-art approaches~\cite{schumann2020tiv,zhang2024tiv}, we are able to work on single-scan radar data and do not rely on scan aggregation, which can increase memory consumption and induce latencies, making it unsuitable for tasks requiring immediate feedback, such as collision avoidance.

The main contribution of this paper is a novel approach for accurate panoptic segmentation in sparse and noisy radar point clouds. Our approach, called SemRaFiner, leverages the advantages of moving instance segmentation, which does not require differentiation between semantic classes but allows us to enhance the overall performance. We optimize our network and propose a new self-attention module, the radius transformer layer, to account for the changing density of radar point clouds, especially when processing moving instance predictions. We utilize dedicated data augmentation to further refine the instance prediction and enhance panoptic segmentation. We construct an efficient network incoporating the individual modules and an optimized training approach to improve the performance of the overall estimation task.

In sum, we make four key claims: First, our SemRaFiner shows state-of-the-art performance for panoptic segmentation in sparse and noisy radar point clouds. 
Second, our radius transformer layer enhances feature extraction, especially for processing the moving object prediction. Third, our optimized training process, including dedicated data augmentation to refine instances and semantic prediction, enhances the overall performance. Fourth, our proposed approach runs faster than the sensor frame rate. 
These claims are backed up by the paper and our experimental evaluation.

\section{Related Work}
\label{sec:related}
The panotic segmentation~\cite{li2022cvpr,sirohi2021tro} unifies semantic segmentation~\cite{chi2022iros,mersch2022ral,sun2022iros} and instance segmentation~\cite{liang2021iccv,schult2022arxiv,vu2022cvpr}. To provide a complete overview, we include different tasks in point cloud processing and structure the extensive related work between projection-based, voxel-based, point-based, transformer-based, and hybrid approaches. 

\textbf{Projection-based} methods~\cite{chen2021ral,sun2022iros,sirohi2021tro} project the point clouds into 2D range images to apply successful convolutional neural networks. These approaches are highly effective and profit from advanced image-based algorithms. However, the range representation results in occlusion artefacts among instances. To account for that, several approaches~\cite{qiu2022tmlr,huang2022eccv} project the point cloud to a polar bird's eye view. Panoptic-PolarNet~\cite{zhou2021cvpr} adopts the representation and incorporates an adaptive point cloud pruning technique and instance augmentation to enhance panoptic segmentation. However, the 2D representation leads to back projection errors and discretization artefacts that harm accuracy~\cite{mersch2022ral}.

\textbf{Voxel-based} approaches~\cite{li2022neurips,qiu2022tmlr} prevent back projection errors by keeping the 3D information of the point clouds. Since point clouds are unevenly distributed, the voxel representation results in empty voxels. To overcome latency issues, state-of-the-art methods~\cite{liang2021iccv,vu2022cvpr} apply submanifold sparse convolutions~\cite{graham2018cvpr} to process occupied voxels and reduce the computational complexity. Further, hybrid approaches~\cite{li2022cvpr,qiu2022tmlr} extend the voxel-based processing and include multiple point cloud representations to alleviate limitations and extract valuable features. Hence, Su~\etalcite{su2023aaai} adopted RPVNet~\cite{xu2021iccv} to include various representations to enhance fine-grained segmentation. Since voxels inherently introduce discretization artefacts, reducing information loss is beneficial, especially in sparse radar data.

\textbf{Point-based} methods~\cite{dubey2022mlwa,qi2017nips2,schumann2018icif} utilize the advantage of directly processing the unordered point cloud data. Qi~\etalcite{qi2017cvpr} propose an innovative approach to use shared multi-layer perceptrons~(MLPs) and max pooling to extract features. PointNet++~\cite{qi2017nips2} extends this approach to capture strong local features by hierarchical feature extraction from larger local regions. Schumann~\etalcite{schumann2018icif} optimize PointNet++ to process sparse radar data. However, for sparse radar data, the processing often relies on aggregated point clouds~\cite{schumann2018icif,schumann2020tiv}, which can be disadvantageous because of the increased latencies and memory consumption. 

Another way of extracting valuable point-wise features within local areas is kernel point convolutions~\cite{chi2022iros,thomas2019iccv}. 
Gasperini~\etalcite{gasperini2021ral} extend the kernel point convolutions backbone with class-dependent post-processing to achieve precise instance segmentation. Nevertheless, the kernel point convolution approaches are often outperformed by attention-based approaches~\cite{zhao2021iccv}, which extract stronger dependency information that is helpful for sparse radar point clouds~\cite{zeller2023ral}.

\textbf{Transformer-based} methods made tremendous progress within recent years and led to major improvements starting from natural language processing~\cite{vaswani2017nips} to point-based scene understanding~\cite{lai2023cvpr,kong2023iccv,xin2022cvpr,zhang2024tiv}. Zhao~\etalcite{zhao2021iccv} propose vector attention~\cite{zhao2020cvpr} on point clouds based on $k$ nearest neighbours to weight the individual feature channels to enable fine-grained feature extraction. PTv2~\cite{wu2022neurips} and PTv3~\cite{wu2024cvpr} take this idea further and propose group vector attention and point cloud serialization, respectively. Lai~\etalcite{xin2022cvpr} utilize a stratified key-sampling strategy within a grid representation to increase the receptive fields of the attention mechanism. Recently, masked-based predictions~\cite{marcuzzi2023ral-meem,schult2022arxiv,xiao2023arxiv} further enhanced the performance. To leverage the potential, several transformer-based approaches~\cite{zeller2023ral,zeller2024tro} are optimized to process sparse and noisy radar data.

In contrast to the related work, we propose a novel method inspired by self-attention that accounts for the sparsity and changing density of radar point clouds. Our SemRaFiner directly utilizes moving instance predictions to leverage the advantage of the radar data to solve the task of panoptic segmentation. We incorporate point-based refinement to extract valuable features and enhance instance segmentation. Furthermore, we optimize the learning process to achieve state-of-the-art performance of panoptic segmentation in sparse and noisy radar data. This makes our approach stand out and different to the existing methods.

\begin{figure*}[t]
 \centering
 \fontsize{8pt}{8pt}\selectfont
 \def\svgwidth{\textwidth}
 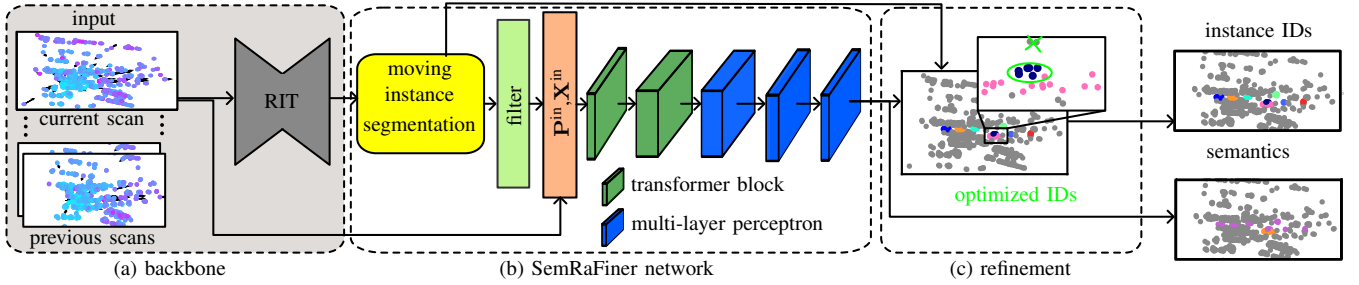%
 \caption{The detailed design of the individual modules of our SemRaFiner. (a) The backbone provides the moving instance predictions. (b) Our network extracts valuable features from filtered moving instances. (c)~The semantic prediction of our approach refines the instance assignment. The final output includes the semantic predictions and corrected instance IDs. The depth of the transformer blocks and multi-layer perceptrons correspond to the feature channels. }
 \vspace{-0.1cm}
 \label{fig:modules}
  \vspace{-0.2cm}
\end{figure*}

\section{Our Approach to Panoptic Segmentation}
\label{sec:main}
The goal of our approach is to achieve precise panoptic segmentation in sparse and noisy radar point clouds to enhance scene understanding of autonomous vehicles. Our SemRaFiner architecture, illustrated in \figref{fig:modules}, utilizes the prediction of the Radar Instance Transformer~\cite{zeller2024icra,zeller2024tro} and extends the backbone to refine instance assignments and include semantic classes. Our network is a point-based transformer network that builds upon the self-attention mechanism~\cite{vaswani2017nips}. We directly process the individual points using our radius transformer layer to enhance feature extraction by addressing the specific spatial distribution of radar points. We further enhance the instance prediction by enabling the network to account for false predictions through our training procedure. The final point-based prediction includes the corresponding class and refined instance assignments.

\subsection{Moving Instance Segmentation Backbone}
\label{sec:misb}
Radar data is typically sparse and noisy, but it provides additional Doppler information that is useful for moving instance segmentation. Due to the enhanced performance of moving instance segmentation~\cite{zeller2024icra,zeller2024tro} compared to standard semantic segmentation~\cite{zeller2023ral}, our idea is to exploit this performance advantage to improve accuracy. Therefore, we utilize the state-of-the-art moving instance segmentation method Radar Instance Transformer~(RIT)~\cite{zeller2024icra,zeller2024tro} to reliably determine moving instances to improve panoptic segmentation. 

RIT enriches the current radar point cloud $\mathcal{P}^{t}$ at time~$t$, consisting of the coordinates of the points and the radar features, including the Doppler velocity and RCS values, with temporal information from $T$ previous scans $\mathcal{P}^{t-T},\dots,\mathcal{P}^{t-1}$ within an optimized module to reduce computational complexity. The enriched point cloud is processed by transformer blocks~\cite{zhao2020cvpr} and attention-based graph clustering to determine the instance IDs ${\mathcal{I}=\{I_1,\dots, I_N\}}$ with \mbox{$I_i\in\NN$}. Additionally, the outputs include the moving object segmentation labels $\mathcal{S}^{\mathrm{MOS}}$. We utilize the predicted labels to filter the point clouds for moving prediction, which is the input to our method. Since we do not rely on extracted features of the backbone, we can potentially substitute the network for other moving instance segmentation approaches. However, the good performance of the RIT is crucial for our method to improve panoptic segmentation. 

\subsection{Radius Transformer Layer}

In sparse radar point clouds, individual points can represent whole instances, such as distant vehicles or nearby pedestrians. Therefore, they contain important information for safe autonomous mobility. Hence, the feature extraction is key to enhancing performance for downstream tasks. Since we process the filtered point cloud, often including only moving instances, the point clouds are sparse and unevenly distributed compared to the unfiltered point cloud, including static points. To address this issue, especially for moving instance prediction, we propose a radius transformer, which combines a ball query sampling strategy with a vector attention mechanism. We argue that the most important information for correctly classifying individual points is within the local area of the object. Most state-of-the-art approaches also focus on the local area. However, due to the filtering of the point clouds explained in~\secref{sec:misb}, individual instances are distributed over the whole field of view.

Performing attention between the $k$ nearest neighbours~\cite{wu2022neurips,zeller2024tro,zhao2021iccv} or determining the corresponding points with serialization algorithms~\cite{wu2024cvpr} can result in interconnections of distant points, which is likely to harm the feature extraction capability for our type of problem. In contrast, our radius-based approach limits the considered related local area. Other approaches~\cite{lai2023cvpr,xin2022cvpr} restrict the local area by performing attention on predefined areas. However, the grid representation is fixed, which does not necessarily align with the instances. Therefore, we propose a radius instance transformer layer, defining the local area for each point individually to overcome this shortcoming.

The input to our transformer layer contains the information about the moving instances within the single current scan $\mathcal{P}^{t}$ at time~$t$. 
This comprises the point coordinates \mbox{$\mathbf{P}^{\text{in}}=[\mathbf{p}_1, \dots, \mathbf{p}_{{N}^{\text{mov}}}]^{\top} \in \RR^{{N}^{\text{mov}}\times 2}$} and the encoded point-wise features $\mathbf{X}^{\text{in}}=[\mathbf{x}_1,\dots,\mathbf{x}_{{N}^{\text{mov}}}]^{\top} \in \RR^{{N}^{\text{mov}}\times D_1}$, where $\mathbf{p}_i\in\RR^{2}$ and $\mathbf{x}_i\in\RR^{D_1}$ for ${N}^{\text{mov}}$ moving points. The features $\mathbf{X}^{\text{in}}$ are the original radar features of the subset of points belonging to moving instances. 
During the training, we utilize the ground truth labels to select the corresponding moving points, and during inference, the points correspond to the semantic and instance predictions of the backbone. We do not process the static predictions. In our transformer layer, we first encode the features of the moving points $\mathbf{X}^{\text{in}}$ as keys $\mathbf{K}$, queries $\mathbf{Q}$, and values $\mathbf{V}$ as follows:
\begin{align}
\mathbf{Q} &= \mathbf{X}^{\text{in}} \mathbf{W}_Q\text{,} &
\mathbf{K} &= \mathbf{X}^{\text{in}} \mathbf{W}_K\text{,} &
\mathbf{V} &= \mathbf{X}^{\text{in}} \mathbf{W}_V\text{,}
\label{eq:1}
\end{align}
where $\mathbf{W}_{K}$, $\mathbf{W}_{Q}$, and $\mathbf{W}_{V} \in \RR^{D_1\times D_1}$ are learned linear projections. 
To adaptively combine the features within the local areas and extract valuable features for individual points, we need to group the points. As mentioned before, to account for the changing density of points and replace the grid representation with a flexible solution, we sample the points within a circle defined by the radius $r$, around the individual points. We determine the neighbourhood of the corresponding points with point coordinates $\mathbf{p}$ by the relative position $\mathbf{r}_{i,j}=\mathbf{p}_i-\mathbf{p}_j$ where $\mathbf{p}_i$ and $\mathbf{p}_j \in \mathbf{P}^{\text{in}}$. The domain of definition for our self-attention is then the circle \mbox{$\mathcal{B}_r(\mathbf{p}_i) = \left\lbrace \mathbf{p}_i \in \mathbf{P}^{\text{in}} \: | \: \left\Vert \mathbf{r}_{i,j} \right\Vert \le r \right\rbrace$} with radius $r$, similar to the ball query of PointNet++~\cite{qi2017nips2}. However, we replace the processing within the local areas of PointNet++ with a powerful attention mechanism. Furthermore, we observe that having many interconnections between points can make it difficult to extract valuable features within local regions, consisting of different semantic classes, due to the information exchange within the attention mechanism. Hence, we restrict the maximum number of points ${N}^{\text{max}}$ within the radius, resulting in an optimized neighbourhood representation to enhance feature extraction.

We sample the points using a ball query~\cite{thomas2019iccv} to extract the related queries, keys, and values, resulting in $\mathbf{Q}^{\text{r}},\mathbf{K}^{\text{r}}$, and $\mathbf{V}^{\text{r}} \in \RR^{N^{\text{mov}} \times N^{\text{max}} \times D_1}$ within the local neighbourhoods. If the local area around the points contains less than ${N}^{\text{max}}$ points, the remaining keys, queries, and values are zero-padded. Furthermore, we utilize the relative position $\mathbf{r}_{i,j}$, which we calculate to determine the local areas to calculate the relative positional encoding.
We process the relative positions $\mathbf{r}_{i,j}$ by two linear layers with weight matrices $\mathbf{W}^p_1 \in \RR^{2\times 2}$ and $\mathbf{W}^p_2 \in \RR^{2\times D_1}$, batch normalization~\cite{ioffe2015icml}, and the ReLU activation function~\cite{nair2010icml} to determine the relative positional encoding $\mathbf{R} \in \RR^{N^{\text{mov}} \times N^{\text{max}} \times D_1}$~\cite{zhao2021iccv}. We adopt vector attention~\cite{zhao2020cvpr} and subtract the keys from the queries, to calculate the attention weights $\mathbf{A} \in \RR^{N^{\text{mov}} \times N^{\text{max}} \times D_1}$ as follows:
\begin{align}
\mathbf{A}_{i,j}= (\mathbf{Q}^{\text{r}}_{i,j}-\mathbf{K}^{\text{r}}_{i,j})+\mathbf{R}_{i,j},
\label{eq:4}
\end{align}
where we add the relative positional encoding $\mathbf{R}$ to include fine-grained position information in the weighting. We process the resulting attention weights by a multi-layer perceptron~(MLP), including two linear layers with weight matrices $\mathbf{W}_1,\mathbf{W}_2 \in \RR^{D_1\times D_1}$, two batch normalization~\cite{ioffe2015icml} layers, and ReLU activation~\cite{nair2010icml}. We then apply the softmax function to derive the final attention weights~$\hat{\mathbf{A}}_{i,j}$.
 The output features $\mathbf{X}^{\text{out}} \in \RR^{N^{\text{mov}} \times D_1}$ of the transformer layer are the sum of the element-wise multiplication, indicated by $\odot$, and the relative positional encoding, determined as follows:
\begin{align}
\mathbf{X}^{\text{out}}_{i} &= \sum_{j=1}^{N^{\text{max}}}{ \hat{\mathbf{A}}_{i,j}\odot (\mathbf{V}^{\text{r}}_{i,j}+\mathbf{R}_{i,j}) }.
\label{eq:5}
\end{align}

Besides the features, which comprise the information within the local area, the output of the transformer layer includes the point coordinates $\mathbf{P}^{\text{in}}$ of the moving points. The point coordinates are kept as is to include fine-grained position information within the consecutive layers.
\begin{figure}[t]
  \centering
  \fontsize{8pt}{8pt}\selectfont
     \def\svgwidth{\linewidth}
     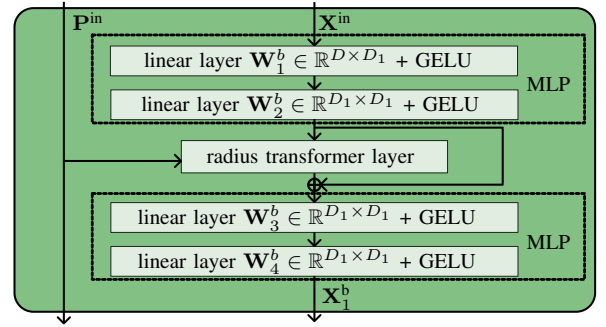
  \caption{Our transformer block is a residual block that includes our radius transformer layer and two multi-layer perceptrons with two linear layers and GELU activation functions. The input includes the point coordinates $\mathbf{P}^{\text{in}}$ and the point-wise features $\mathbf{X}^{\text{in}}$.}
  \label{fig:transformerblock}
  \vspace{-0.4cm}
\end{figure}
\subsection{SemRaFiner Network}
We aim for an efficient processing of the radar data to reduce latencies. Consequently, our SemRaFiner includes only two transformer blocks and three multi-layer perceptrons, to derive the refined instance IDs $\mathcal{I}^r=\{I^r_1,\dots, I^r_N\}$ with $I^r_i\in\NN$ and semantic labels $\mathcal{S}^{\mathrm{sem}}$, see~\figref{fig:modules}. 

We embed our radius transformer layer into a residual transformer block to extract valuable features, see~\figref{fig:transformerblock}. We first increase the dimension of the features $\mathbf{X}^{\text{in}}$, which we then process within the radius transformer layer. The output features of the radius transformer layer are added to the features of the skip connection and processed by the second MLP. The point coordinates are directly passed into the transformer layer to calculate the positional encoding to include fine-grained position information within the local areas. The resulting output includes the point-wise features and coordinates to keep position information intact. The resulting feature after the second transformer block $\mathbf{X}_{2}^{\text{b}} \in \RR^{N^{\text{mov}} \times D_2}$ are processed by the three consecutive MLPs, illustrated by the blue blocks in~\figref{fig:modules}, to reduce the dimensionality and derive the final predictions. The first two MLPs consist of two linear layers, where the first linear layer keeps the dimension of the features, and the second reduces the dimension by a factor of two. To derive the final semantic predictions, the last MLP first reduces the dimensionality by a factor of two before predicting the semantic labels. The final semantic prediction $\mathcal{S}^{\mathrm{sem}}=\{s^{\mathrm{sem}}_1,\dots,s^{\mathrm{sem}}_{N}\}$ where $s^{\mathrm{sem}}_i\in \{1, \dots, C\}$ combine the static predictions of the backbone and the refined semantic class predictions. 

\subsection{Data Augmentation}
\label{sec:data}
We utilize ground truth annotations to train our model to classify instances within sparse radar point clouds reliably. During inference, we leverage the moving instance prediction to perform point-wise semantic segmentation and refine the instance prediction. The major problem is that the ground truth does not include failure cases such as static points predicted as moving. To account for the false prediction of static points, we augment our training data. Our idea is to enable our network to correct false predictions, which cannot be learnt by only considering the ground truth annotations. Therefore, we incorporate static points into the training process and add static points close to an instance by the probability of $p_I$. This approach also aligns with the false predictions, which often occur at the boundary of the instance. Additionally, we observe that the moving instance predictions include false predictions for small instances within the individual scans. Therefore, we add instances comprising only static ground truth annotations, including one to five points by the probability of $p_S$. 
\subsection{Instance Refinement}
\label{sec:ref}
Our data augmentation enables us to correct false predictions and remove static points from instance predictions. Furthermore, we utilize the semantic predictions to correct instance assignments. Since the classification of moving and static points does not include further differentiation, we observe that nearby instances belonging to different semantic classes, such as trucks and cars, may get clustered together. However, our panoptic segmentation incorporates that knowledge, and hence, we can resolve the false assignments. We combine the instance assignments and semantic predictions within our instance refinement module. Due to the point-wise processing, we derive a semantic class for each individual radar point. We combine these predictions and the instance assignment per point and assign different instance IDs if the semantic classes differ. The overall refinement thus accounts for false semantic predictions and instance assignments. 
\begin{table*}[t]
\resizebox{\textwidth}{!}{
\begin{tabular}{l|ccc|cccccc|cccccc}
\hline
                           & \multicolumn{1}{l}{}                                  &                                &                                & \multicolumn{6}{c|}{PQ}                                                                                                                                                                            & \multicolumn{6}{c}{IoU}                                                                                                                                                                              \\ \cline{5-16} 
\multirow{-2}{*}{Method}   & \multicolumn{1}{l}{\multirow{-2}{*}{Input}}           & \multirow{-2}{*}{PQ}         & \multirow{-2}{*}{mIoU}           & static                         & car                            & ped.                           & ped. grp.                      & bike                           & truck                          & static                         & car                            & ped.                           & ped. grp.                      & bike                           & truck                          \\ \hline
\rowcolor[HTML]{E0E0E0} 
RadarPNv1~\cite{schumann2018icif}                  & \cellcolor[HTML]{E0E0E0}                              & -                           & 61.0                           & -                           & -                           & -                          & -                          & -                           & -                           & 98.7                           & 58.2                           & 36.0                           & 58.7                           & 58.4                           & 56.1                          \\
\rowcolor[HTML]{E0E0E0} 
RadarPNv2~\cite{schumann2020tiv}                  & \multirow{-2}{*}{\cellcolor[HTML]{E0E0E0}} & -                          & 61.9                         & -                           & -                           & - & -                           & -                           & -                           &  98.7                           & 63.8                           & 38.8& 58.5                           & 51.0                           & 61.0         \\        
\rowcolor[HTML]{E0E0E0} 
STA-Net~\cite{zhang2024tiv}                  & \multirow{-3}{*}{\cellcolor[HTML]{E0E0E0}multi-scan}& -                          & \textbf{70.4}                         & -                           & -                           & - & -                           & -                           & -                          &  \textbf{99.8}                           & \textbf{75.9}                         & \textbf{43.1} & \textbf{82.0}                           & \textbf{67.1}                           & 54.6                            \\ \hline
Mask3D~\cite{schult2022arxiv}                         &                                                       & 56.9                          & 56.1                          & 98.4                           & 70.6                           &18.4                           & 31.9                          & 57.7                          & 64.3                          & 98.6                         & 68.7                          & 11.8                          & 35.9                          & 57.7                          & 63.9                           \\
GRT~\cite{zeller2023ral} + DBSCAN~\cite{ester1996kdd2}         &                                                       & 56.3                           & 56.9                           & 98.7                          & 59.0                          & 29.7                           & 38.3                           & 54.6                           & 57.3                           & 98.7                           & 58.5                           & 21.7                           & 46.0                           & 54.6                           & 61.8                           \\
Ours &        \multirow{-3}{*}{single-scan}                                        & \textbf{81.4} & \textbf{70.4} & \textbf{99.7} & \textbf{85.7} & \textbf{59.2}                           & \textbf{83.1} & \textbf{78.9} & \textbf{82.1} & 99.4 & 74.9 & 42.1                          & 71.0 & 65.2 & \textbf{69.6}\\ \hline
\end{tabular}%
}
\caption{Panoptic segmentation results on the RadarScenes test set in terms of $\mathrm{PQ}$ and $\mathrm{IoU}$ scores. The results of RadarPNv1~\cite{schumann2018icif}, RadarPNv2~\cite{schumann2020tiv} and STA-Net~\cite{zhang2024tiv} are adapted from the original papers. The gray rows correspond to multi-scan segmentation methods.
  \vspace{-0.4cm}}
  \label{tab:resall}
\end{table*}
\subsection{Implementation Details}
\label{sec:impl}
We train the backbone separately and adopt the original parameters~\cite{zeller2024tro}. We implemented our SemRaFiner in PyTorch~\cite{paszke2019nips} and trained our network with one Nvidia A100 GPU and a batch size of 64 over 80 epochs. As optimizer, we utilize AdamW~\cite{loshchilov2017iclr}, set the initial learning rate to 0.001 and drop the learning rate by a factor of 10 after 60 epochs. We combine the Lovász loss~\cite{berman2018cvpr} and cross-entropy loss to learn the point-wise classification using the same weighting for all classes. 

Additionally, we introduce a consistency loss to enforce the same class prediction for the individual instances. We construct a loss based on the semantic predictions $\mathcal{I}_h^{\mathrm{sem}}$ for the individual instances $h$. The ideal case is that each instance belongs to the same class. However, this limits the performance since we cannot correct false predictions during instance refinement. Therefore, it is important to maintain point-wise classification while preserving consistency within the true instances. We utilize the number of distinct classes within an instance $\lvert \mathcal{I}_h^{\mathrm{sem}}\rvert$ to calculate the loss over the number of all instances~$N_h$ as follows: 
\begin{align}
\mathcal{L}_c= \frac{1}{N_h}\sum^{N_h}_{h = 1} 1- \frac{1}{\lvert \mathcal{I}_h^{\mathrm{sem}}\rvert}. 
\label{eq:1loss}
\end{align}

We combine all three losses without weighting to derive the final training loss for our network.

We process the input features with dimension $D=5$, comprising the point coordinates $(x^C_i, y^C_i,z^C_i)$, the radar cross section $\sigma_i$, and the ego-motion-compensated Doppler velocity~$v_i$, resulting in a five-dimensional input vector $\mathbf{x}_i=(x^C_i, y^C_i, z^C_i, \sigma_i, v_i)$. The coordinate $z^C_{i}=0$ is added to apply the pose compensation. We increase the 
per-point features to $D_1=64$, $D_2=256$ within our transformer blocks. We set the parameters for our radius transformer layer to $r=\SI{5.0}{m}$ and the maximum considered neighbours within the ball query to ${N}^{\text{max}}=24$. The first two MLPs reduce the dimensionality to $128$ and $64$, respectively. In the third MLP, the first linear layer has an output dimension of $32$ before predicting $C$ classes by the second linear layer. We set the probabilities of augmenting the instances $p_I$ and the scan $p_S$ to $40\,\%$.

\section{Experimental Evaluation}
\label{sec:exp}

The main focus of this work is to achieve accurate panoptic segmentation in sparse radar point clouds.
We present our experiments to show the capabilities of our method. The results of our experiments support our key claims, including that our approach achieves state-of-the-art performance in panoptic segmentation without aggregating scans. Moreover, our radius instance transformer layer accounts for the changing density in sparse radar data to enhance panoptic segmentation. Our dedicated data augmentation and point-wise processing enables instance refinement to improve overall performance. Our SemRaFiner is efficient and only adds a small overhead to the backbone to incorporate semantic classes. 

\subsection{Experimental Setup}
We train and evaluate our model on the RadarScenes~\cite{schumann2021icif} dataset since it is the only large-scale open-source radar dataset~\cite{zhou2022sensors} that includes per-point annotations for moving instance segmentation and panoptic segmentation under different weather conditions and driving scenarios. As described by Zeller~\etalcite{zeller2024tro}, we split the 158 sequences into 130 sequences for training, 6 for validation, and 22 for testing. We perform all ablation studies on the validation set. We adopt the sparse representation of the single scan data processing~\cite{zeller2024tro} containing up to four radar sensors to cover the surroundings of the vehicle.

We utilize the panoptic quality~($\mathrm{PQ}$)~\cite{behley2021icra} to evaluate the panoptic segmentation. Additionally, we report the mean intersection over union~($\mathrm{mIoU})$ to evaluate the semantic segmentation performance in detail. To enable a detailed evaluation, we further differentiate between the six classes, including static, pedestrian~(ped.), pedestrian group~(ped. grp.), car, truck, and bike. 

\subsection{Panoptic Segmentation}
\label{sec:mit}
The first experiment evaluates the performance of our approach and its outcomes
support the claim that our approach achieves state-of-the-art performance in panoptic segmentation in sparse and noisy radar scans. We compare our method to high-performing networks with strong performance in point-based instance segmentation and radar segmentation. We furthermore utilize Mask3D~\cite{schult2022arxiv} as a baseline and extend Gaussian Radar Transformer~(GRT)~\cite{zeller2023ral} with DBSCAN~\cite{ester1996kdd2} to reliably cluster instances. We extend Mask3D~\cite{schult2022arxiv} with state-of-the-art post-processing for mask predictions~\cite{cheng2022cvpr} and set the score threshold to $0.5$ to enhance performance and derive the panoptic labels. Additionally, we report the segmentation results of RadarPNv1~\cite{schumann2018icif}, RadarPNv2~\cite{schumann2020tiv}, and STA-Net~\cite{zhang2024tiv}, which rely on scan aggregation and do not consider single-scan processing. 

Our SemRaFiner outperforms the existing single-scan approaches, especially in terms of $\mathrm{PQ}$, by a solid margin, as displayed in~\tabref{tab:resall}. The performance improves, especially for small instances comprising only a few points such as the pedestrians. We observe that our dedicated network, which utilizes the moving instance segmentation as input, is able to extract valuable features, particularly for smaller classes, due to the reduced class imbalance and optimized architecture. We emphasize that our performance depends on the prediction of the RIT, and therefore, precise moving instance segmentation is important to enhance the accuracy. 

Another way to enhance segmentation performance is by aggregating radar point clouds. However, the aggregation of point clouds can induce disadvantageous latencies, which is relevant when performing our approach on a vehicle in real time. Additionally, our method performs on par in terms of $\mathrm{mIoU}$ with the best-performing semantic segmentation model STA-Net, which aggregates scans over $500\,\mathrm{ms}$ with input point clouds of $3072$ points. In contrast, our single scans include an average number of points of $539$ in the test set. The RIT utilizes two previous scans to enrich the current scan, but only the single current scan is processed within the backbone. Therefore, we achieve similar results using five times fewer points and also include instance segmentation.

GRT achieves a higher $\mathrm{mIoU}$ compared to Mask3D for the single-scan approaches. We assume that the optimized radar-specific architecture performs better on radar data. However, the powerful mask predictions lead to better performance in terms of $\mathrm{PQ}$ but come with a higher computational cost. Overall, both methods struggle to reliably segment instances in sparse and noisy radar point clouds. We enhance the $\mathrm{PQ}$ for all classes, leading to an improvement of more than 20 percentage points on the overall $\mathrm{PQ}$.
\begin{table}[t]
  \centering
  \setlength\tabcolsep{3pt}
{%
\begin{tabular}{@{}cl|cc|cc@{}}
\toprule
\#&Model                      &$r\,[m]$&$N^{max}$ & $\mathrm{PQ}$            & $\mathrm{mIoU}$             \\ \midrule
{[}A] &Stratified Transformer~\cite{lai2023cvpr}     & & & 55.0    & 42.6\\
{[}B]& Point Transformer~\cite{zhao2020cvpr}  & &&  75.8  & 63.1\\
{[}C] &KPConv~\cite{thomas2019iccv}     & & & 81.0    & 69.6 \\
{[}D]& Ours                     & 5.0 & 24 & \textbf{83.0}  & \textbf{72.2}       \\ \bottomrule
{[}E]& Ours                     &6.0 & 24 & 82.3  & 71.3 \\
{[}F]& Ours                     &6.0 & 34 & 82.4 & 71.5 \\
{[}G] &Ours                     & 5.0& 34 &82.7  & 72.0 \\
{[}H]& Ours                     & 3.0& 24 & 81.7  & 70.4 \\
{[}I]& Ours                     & 5.0&  6& 78.7 & 66.4 \\ \bottomrule
\end{tabular}%
}
\caption{Ablation study results on the RadarScenes validation set in terms of $\mathrm{PQ}$ and $\mathrm{mIoU}$ scores.}
  \label{tab:radius}
  \vspace{-0.4cm}
\end{table}
\begin{figure}[t]
  \centering
  \fontsize{8pt}{8pt}\selectfont
     \def\svgwidth{\linewidth}
     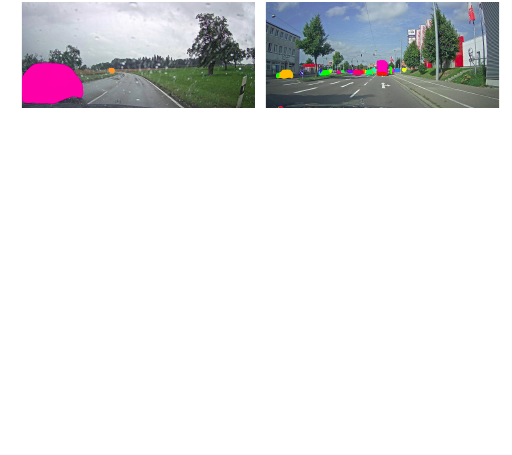
  \caption{Qualitative results of our SemRaFiner and the Radar Instance Transformer~(RIT)~\cite{zeller2024tro} on the RadarScenes dataset. The camera images are anonymized and just shown for reference. The left column is from sequence 79 (fog), and the right is from sequence 107 (urban environment). Each colour in the image of the predictions represents a different instance (static is grey).}
  \vspace{-0.2cm}
  \label{fig:qres}
  \vspace{-0.4cm}
\end{figure}
\subsection{Ablation Study on the Radius Transformer Layer}
\label{sec:ablrad}
The second experiment evaluates our radius transformer layer and illustrates that our approach is capable of extracting valuable features from sparse point clouds, enhancing the overall performance. We replace the radius transformer layer with the point transformer layer~\cite{zhao2020cvpr} used for radar signal processing~\cite{zeller2024tro}, the Stratified transformer layer~\cite{lai2023cvpr}, and KPConv~\cite{thomas2019iccv}, illustrating the ablation results in~\tabref{tab:radius}. Additionally, we evaluate the performance of our approach based on the selection of the hyperparameters, namely the radius $r$ and the maximum number of neighbours $N^{max}$. 

The gain obtained by our radius transformer layer is directly visible by the increase compared to the other transformer modules, enhancing the performance by more than 6 absolute percentage points compared to the point transformer layer. We argue that the $k$NN-based processing results in the interconnection of distant points, and the grid representation does not align with the instances, both harming the accuracy. As expected, the KPConv approach,~\tabref{tab:radius} [C], leads to the best result using the ball query to determine the local areas. However, our attention-based processing enhances the $PQ$ by 2 absolute percentage points compared to KPConv. 

 Furthermore, the comparison of the setup of $r=\SI{5}{m}$ and $N^{\mathrm{max}}=6$,~\tabref{tab:radius} [I], which utilizes the same number of neighbours as the 
point transformer, underlines that the additional neighbourhood restrictions by the radius are advantageous. We argue that especially for small instances such as pedestrians but also far away cars or bikes lead to too many interconnections harm the accuracy. However, for larger instances, we observe that increasing the number of neighbours helps to classify them reliably. 

The performance decreases if we increase the maximum number of points from 24 to 34,~\tabref{tab:radius} [D] and [G], respectively. We assume that too many interactions result in a mix of information during self-attention, which hinders the clear separation of the information. We also see that for a larger number of neighbours, the performance of smaller instances, such as pedestrians, drops from 65.2\,\% to 64.0\,\% for $N^{max}=24$ and $N^{max}=34$, respectively. Since close instances might still be included within the radius, the performance of small instances, especially, is affected. Therefore, the radius and the maximum number of parameters depend on the task and the segmentation of the specific instances. Additionally, the parameters may depend on the resolution of the radar sensor, which is an interesting starting point for future research to automatically determine the optimal settings. Overall, both parameters are essential to enhance radar-based panoptic segmentation. 

\begin{table}[t]
  \centering
  \setlength\tabcolsep{3pt}
{%
\begin{tabular}{@{}ll@{}}
\toprule
Model                      &$\mathrm{PQ}$                         \\ \midrule
Ground truth semantics      & 77.1    \\
Ours w/o data augmentation \quad \quad \quad\quad&  81.5 \\
Ours                      & \textbf{83.0}       \\ \bottomrule
\end{tabular}%
}
\caption{Panoptic segmentation result on the RadarScenes validation set in terms of $\mathrm{PQ}$ and $\mathrm{mIoU}$ scores.}
  \label{tab:ref}
  \vspace{-0.2cm}
\end{table}
\begin{table}[t]
  \centering
{%
\begin{tabular}{@{}lrr@{}}
\toprule
Model                     & parameters (M) & mean runtime (ms) \\\midrule
\rowcolor[HTML]{E0E0E0} STA-Net~\cite{zhang2024tiv}                        & 7.4            & 34.9         \\
\rowcolor[HTML]{E0E0E0} RadarPNv1~\cite{schumann2018icif}                       & 1.8           & 278.1         \\ \midrule
Mask3D~\cite{schult2022arxiv}                    & 39.6           & 85.2         \\
Gaussian Radar Transformer~\cite{zeller2023ral}&       8.4         & 24.0         \\
Radar Instance Transformer~\cite{zeller2024tro}                      & 3.8            & 31.7         \\
\midrule
Ours + RIT~\cite{zeller2024tro}                       & 4.5            & 42.1         \\

 \bottomrule
\end{tabular}%
}
\caption{The evaluation of the mean runtime and the number of parameters of the models on an Nvidia RTX A6000 GPU based on 1000 randomly sampled point clouds of the RadarScenes data set. The runtime and the parameters of STA-Net and RadarPNv1 are copied from Zhang~\etalcite{zhang2024tiv} using a different hardware setup. }
  \label{tab:runtimetabl}
  \vspace{-0.6cm}
\end{table}
\subsection{Ablation Study on the Instance Refinement}
\label{sec:ablref}
The third experiment supports the claim that our method incorporates efficient instance refinement to enhance overall performance, as depicted in~\tabref{tab:ref}. The central parts that enable instance refinement are data augmentation, including static points and point-wise processing. To verify the assumptions, we first evaluate the performance of the moving instance segmentation prediction in combination with the semantic ground truth labels. We, therefore, do not optimize the instance prediction of the Radar Instance Transformer~(RIT) to illustrate the limitation of keeping the instance assignments under perfect segmentation results. In the second step, we remove the data augmentation introduced in~\secref{sec:data}. The resulting method accounts for false predictions within the different object classes but cannot correct false static predictions because these errors are excluded in the training process. However, our instance refinement helps to optimize the instance prediction compared to RIT even under the condition that our semantic predictions are not perfect. Furthermore, the data augmentation enhances the panoptic quality by more than two absolute percentage points. Therefore, data augmentation and point-wise processing are essential to improve performance, especially for the annotated moving objects in RadarScenes. Furthermore, we removed the consistency loss for the class consistency within instances, which resulted in a $\mathrm{PQ}$ of $82.9\,\mathrm{\%}$, leading to a small decrease in performance. Furthermore, our SemRaFiner works reliably under versatile scenes, as illustrated in~\figref{fig:qres}. Our approach reliably refines the moving instances under different and changing
scenarios.
\subsection{Runtime}
Finally, we analyze the runtime and the number of parameters of the approaches and, in this way, support the claim that our approach runs fast enough to support online processing. We, therefore, tested our approach on an Nvidia RTX A6000 GPU on 1,000 real-world radar scans that we randomly selected from the RadarScenes validation set. The mean runtime for one radar scan is illustrated in~\tabref{tab:runtimetabl}. The Gaussian Radar Transformer has the lowest runtime utilizing optimized farthest point sampling and $k$NN algorithm in C++~\cite{zhao2021iccv}. The multi-scan approaches increase the runtime compared to the Gaussian Radar Transformer, but the comparison is difficult since the reported results are evaluated on different hardware. Additionally, Mask3D uses four times more parameters to enhance the panoptic segmentation. The Radar Instance Transformer and our approach
combined still have fewer parameters compared to STA-Net since our model only adds a small overhead. The combined runtime of 42.1\,ms, equal to 24\,Hz, is faster than the frame rate of 17\,Hz of the radar sensors.

In summary, our evaluation suggests that our method
provides competitive panoptic segmentation in sparse radar
point clouds. The instance refinement and the utilization of moving instance predictions outperform state-of-the-art approaches. Thus, we support all our claims made in this paper through our experimental evaluation.

\section{Conclusion}
\label{sec:conclusion}

In this paper, we presented a novel approach for panoptic segmentation of sparse and noisy radar point clouds. Our method exploits the advantages of moving instance segmentation to refine the predictions and enhance the overall system performance.
Our method efficiently exploits the sparse representation of moving points by limiting the self-attention mechanism within a local neighbourhood.
We overcome the limitations of grid representations and unbounded local regions in transformer networks to predict the semantic classes. Our point-wise processing enables us to refine instance prediction and correct assignment errors. 
This allows us to successfully identify individual instances and enhance feature extraction by optimizing the training procedure. We implemented and evaluated our approach on the RadarScenes dataset and provided comparisons to other existing techniques. The experiments suggest that our proposed modules and training strategy are essential to achieve good performance on panoptic segmentation, supporting all claims made in this paper.
Overall, our approach outperforms the state-of-the-art approaches, taking a step forward towards sensor redundancy for semantic segmentation for autonomous robots and vehicles.

\bibliographystyle{plain_abbrv}

\bibliography{glorified,new}

\end{document}

%% file: stachnisslab-latex.tex
\usepackage{graphics}           
\usepackage{times}              
\usepackage{amsmath}            
\usepackage{amssymb}            
\usepackage{graphicx}
\usepackage{algorithm}
\usepackage[noend]{algpseudocode}
\usepackage{booktabs}
\usepackage{color}

\usepackage[nocompress]{cite} %
\definecolor{instructioncolor}{rgb}{.5,.5,.5}

\usepackage[font=small]{caption}

\def\secref#1{Sec.~\ref{#1}}
\def\figref#1{Fig.~\ref{#1}}
\def\tabref#1{Tab.~\ref{#1}}
\def\eqref#1{Eq.~(\ref{#1})}

\makeatletter
\usepackage{xspace}
\DeclareRobustCommand\onedot{\futurelet\@let@token\@onedot}
\def\@onedot{\ifx\@let@token.\else.\null\fi\xspace}

\def\etal{{et al}\onedot}
\makeatother

\def\etalcite#1{\etal~\cite{#1}}

\usepackage{array}
\newcolumntype{L}[1]{>{\raggedright\let\newline\\\arraybackslash\hspace{0pt}}m{#1}}
\newcolumntype{C}[1]{>{\centering\let\newline\\\arraybackslash\hspace{0pt}}m{#1}}
\newcolumntype{R}[1]{>{\raggedleft\let\newline\\\arraybackslash\hspace{0pt}}m{#1}}

%% file: stachnisslab-math.tex
\newcommand{\RR}{\mathbb{R}}
\newcommand{\NN}{\mathbb{N}}

%% file: pics/motivation_semrafiner.pdf_tex
\begingroup%
  \makeatletter%
  \providecommand\color[2][]{%
    \errmessage{(Inkscape) Color is used for the text in Inkscape, but the package 'color.sty' is not loaded}%
    \renewcommand\color[2][]{}%
  }%
  \providecommand\transparent[1]{%
    \errmessage{(Inkscape) Transparency is used (non-zero) for the text in Inkscape, but the package 'transparent.sty' is not loaded}%
    \renewcommand\transparent[1]{}%
  }%
  \providecommand\rotatebox[2]{#2}%
  \newcommand*\fsize{\dimexpr\f@size pt\relax}%
  \newcommand*\lineheight[1]{\fontsize{\fsize}{#1\fsize}\selectfont}%
  \ifx\svgwidth\undefined%
    \setlength{\unitlength}{245.71800232bp}%
    \ifx\svgscale\undefined%
      \relax%
    \else%
      \setlength{\unitlength}{\unitlength * \real{\svgscale}}%
    \fi%
  \else%
    \setlength{\unitlength}{\svgwidth}%
  \fi%
  \global\let\svgwidth\undefined%
  \global\let\svgscale\undefined%
  \makeatother%
  \begin{picture}(1,0.98079912)%
    \lineheight{1}%
    \setlength\tabcolsep{0pt}%
    \put(0,0){\includegraphics[width=\unitlength,page=1]{pics/motivation_semrafiner.pdf}}%
    \put(0.81863867,0.25753758){\color[rgb]{0,0,0}\makebox(0,0)[lt]{\lineheight{1.25}\smash{\begin{tabular}[t]{l}instances\end{tabular}}}}%
    \put(0.1616608,0.05588205){\color[rgb]{0,0,0}\rotatebox{90}{\makebox(0,0)[lt]{\lineheight{1.25}\smash{\begin{tabular}[t]{l}predictions\end{tabular}}}}}%
    \put(0.1616608,0.35557319){\color[rgb]{0,0,0}\rotatebox{90}{\makebox(0,0)[lt]{\lineheight{1.25}\smash{\begin{tabular}[t]{l}predictions\end{tabular}}}}}%
    \put(0.10784133,0.07632568){\color[rgb]{0,0,0}\rotatebox{90}{\makebox(0,0)[lt]{\lineheight{1.25}\smash{\begin{tabular}[t]{l}\textbf{instance}\end{tabular}}}}}%
    \put(0.05907805,0.08855505){\color[rgb]{0,0,0}\rotatebox{90}{\makebox(0,0)[lt]{\lineheight{1.25}\smash{\begin{tabular}[t]{l}\textbf{refined}\end{tabular}}}}}%
    \put(0.1109911,0.76574657){\color[rgb]{0,0,0}\rotatebox{90}{\makebox(0,0)[lt]{\lineheight{1.25}\smash{\begin{tabular}[t]{l}instance \end{tabular}}}}}%
    \put(0.1616608,0.72655928){\color[rgb]{0,0,0}\rotatebox{90}{\makebox(0,0)[lt]{\lineheight{1.25}\smash{\begin{tabular}[t]{l}segmentation\end{tabular}}}}}%
    \put(0.05788605,0.77288667){\color[rgb]{0,0,0}\rotatebox{90}{\makebox(0,0)[lt]{\lineheight{1.25}\smash{\begin{tabular}[t]{l}moving\end{tabular}}}}}%
    \put(0,0){\includegraphics[width=\unitlength,page=2]{pics/motivation_semrafiner.pdf}}%
    \put(0.81639422,0.86701761){\color[rgb]{0,0,0}\makebox(0,0)[lt]{\lineheight{1.25}\smash{\begin{tabular}[t]{l}instances\end{tabular}}}}%
    \put(0,0){\includegraphics[width=\unitlength,page=3]{pics/motivation_semrafiner.pdf}}%
    \put(0.70349397,0.70100238){\color[rgb]{0,0,0}\makebox(0,0)[lt]{\lineheight{1.25}\smash{\begin{tabular}[t]{l}$\SI{75}{m}$\end{tabular}}}}%
    \put(0,0){\includegraphics[width=\unitlength,page=4]{pics/motivation_semrafiner.pdf}}%
    \put(0.8198765,0.55242936){\color[rgb]{0,0,0}\makebox(0,0)[lt]{\lineheight{1.25}\smash{\begin{tabular}[t]{l}truck\end{tabular}}}}%
    \put(0,0){\includegraphics[width=\unitlength,page=5]{pics/motivation_semrafiner.pdf}}%
    \put(0.81868408,0.52004117){\color[rgb]{0,0,0}\makebox(0,0)[lt]{\lineheight{1.25}\smash{\begin{tabular}[t]{l}car\end{tabular}}}}%
    \put(0.05781422,0.37301788){\color[rgb]{0,0,0}\rotatebox{90}{\makebox(0,0)[lt]{\lineheight{1.25}\smash{\begin{tabular}[t]{l}\textbf{semantic}\end{tabular}}}}}%
    \put(0.10871606,0.32247371){\color[rgb]{0,0,0}\rotatebox{90}{\makebox(0,0)[lt]{\lineheight{1.25}\smash{\begin{tabular}[t]{l}segmentation\end{tabular}}}}}%
    \put(0.70824356,0.31410138){\color[rgb]{0,0,0}\makebox(0,0)[lt]{\lineheight{1.25}\smash{\begin{tabular}[t]{l}$\SI{75}{m}$\end{tabular}}}}%
    \put(0,0){\includegraphics[width=\unitlength,page=6]{pics/motivation_semrafiner.pdf}}%
    \put(0.70651423,0.015809){\color[rgb]{0,0,0}\makebox(0,0)[lt]{\lineheight{1.25}\smash{\begin{tabular}[t]{l}$\SI{75}{m}$\end{tabular}}}}%
    \put(0,0){\includegraphics[width=\unitlength,page=7]{pics/motivation_semrafiner.pdf}}%
    \put(0.44220264,0.61957451){\color[rgb]{0,0,0.50196078}\makebox(0,0)[lt]{\lineheight{1.25}\smash{\begin{tabular}[t]{l}SemRaFiner\end{tabular}}}}%
    \put(0.55687959,0.18715496){\color[rgb]{0,0,0}\makebox(0,0)[lt]{\lineheight{1.25}\smash{\begin{tabular}[t]{l}\textbf{refined}\end{tabular}}}}%
    \put(0,0){\includegraphics[width=\unitlength,page=8]{pics/motivation_semrafiner.pdf}}%
    \put(0.8168306,0.93832494){\color[rgb]{0,0,0}\makebox(0,0)[lt]{\lineheight{1.25}\smash{\begin{tabular}[t]{l}static\end{tabular}}}}%
    \put(0,0){\includegraphics[width=\unitlength,page=9]{pics/motivation_semrafiner.pdf}}%
    \put(0.81563827,0.90593674){\color[rgb]{0,0,0}\makebox(0,0)[lt]{\lineheight{1.25}\smash{\begin{tabular}[t]{l}moving\end{tabular}}}}%
    \put(0,0){\includegraphics[width=\unitlength,page=10]{pics/motivation_semrafiner.pdf}}%
  \end{picture}%
\endgroup%

%% file: pics/semrafiner_architecture.pdf_tex
\begingroup%
  \makeatletter%
  \providecommand\color[2][]{%
    \errmessage{(Inkscape) Color is used for the text in Inkscape, but the package 'color.sty' is not loaded}%
    \renewcommand\color[2][]{}%
  }%
  \providecommand\transparent[1]{%
    \errmessage{(Inkscape) Transparency is used (non-zero) for the text in Inkscape, but the package 'transparent.sty' is not loaded}%
    \renewcommand\transparent[1]{}%
  }%
  \providecommand\rotatebox[2]{#2}%
  \newcommand*\fsize{\dimexpr\f@size pt\relax}%
  \newcommand*\lineheight[1]{\fontsize{\fsize}{#1\fsize}\selectfont}%
  \ifx\svgwidth\undefined%
    \setlength{\unitlength}{505.89001465bp}%
    \ifx\svgscale\undefined%
      \relax%
    \else%
      \setlength{\unitlength}{\unitlength * \real{\svgscale}}%
    \fi%
  \else%
    \setlength{\unitlength}{\svgwidth}%
  \fi%
  \global\let\svgwidth\undefined%
  \global\let\svgscale\undefined%
  \makeatother%
  \begin{picture}(1,0.207555)%
    \lineheight{1}%
    \setlength\tabcolsep{0pt}%
    \put(0,0){\includegraphics[width=\unitlength,page=1]{pics/semrafiner_architecture.pdf}}%
    \put(0.02777558,0.02856643){\color[rgb]{0,0,0}\makebox(0,0)[lt]{\lineheight{1.25}\smash{\begin{tabular}[t]{l}previous scans\end{tabular}}}}%
    \put(0.03486521,0.11591308){\color[rgb]{0,0,0}\makebox(0,0)[lt]{\lineheight{1.25}\smash{\begin{tabular}[t]{l}current scan\end{tabular}}}}%
    \put(0.89046129,0.09116766){\color[rgb]{0,0,0}\makebox(0,0)[lt]{\lineheight{1.25}\smash{\begin{tabular}[t]{l}semantics\end{tabular}}}}%
    \put(0.88945914,0.18008425){\color[rgb]{0,0,0}\makebox(0,0)[lt]{\lineheight{1.25}\smash{\begin{tabular}[t]{l}instance IDs\end{tabular}}}}%
    \put(0.05892939,0.18896395){\color[rgb]{0,0,0}\makebox(0,0)[lt]{\lineheight{1.25}\smash{\begin{tabular}[t]{l}input\end{tabular}}}}%
    \put(0,0){\includegraphics[width=\unitlength,page=2]{pics/semrafiner_architecture.pdf}}%
    \put(0.46867698,0.06723459){\color[rgb]{0,0,0}\makebox(0,0)[lt]{\lineheight{1.25}\smash{\begin{tabular}[t]{l}transformer block\end{tabular}}}}%
    \put(0.46877398,0.0382059){\color[rgb]{0,0,0}\makebox(0,0)[lt]{\lineheight{1.25}\smash{\begin{tabular}[t]{l}multi-layer perceptron\end{tabular}}}}%
    \put(0,0){\includegraphics[width=\unitlength,page=3]{pics/semrafiner_architecture.pdf}}%
    \put(0.08933666,0.00542573){\color[rgb]{0,0,0}\makebox(0,0)[lt]{\lineheight{1.25}\smash{\begin{tabular}[t]{l}(a)  backbone\end{tabular}}}}%
    \put(0.37213887,0.00542573){\color[rgb]{0,0,0}\makebox(0,0)[lt]{\lineheight{1.25}\smash{\begin{tabular}[t]{l}(b)  SemRaFiner network\end{tabular}}}}%
    \put(0.70213518,0.00542573){\color[rgb]{0,0,0}\makebox(0,0)[lt]{\lineheight{1.25}\smash{\begin{tabular}[t]{l}(c)  refinement\end{tabular}}}}%
    \put(0.70590417,0.0599871){\color[rgb]{0,1,0}\makebox(0,0)[lt]{\lineheight{1.25}\smash{\begin{tabular}[t]{l}optimized IDs\end{tabular}}}}%
    \put(0,0){\includegraphics[width=\unitlength,page=4]{pics/semrafiner_architecture.pdf}}%
    \put(0.42283476,0.10658774){\color[rgb]{0,0,0}\rotatebox{90}{\makebox(0,0)[lt]{\lineheight{1.25}\smash{\begin{tabular}[t]{l}$\mathbf{P}^{\text{in}}$,$\mathbf{X}^{\text{in}}$\end{tabular}}}}}%
    \put(0.38796856,0.11520546){\color[rgb]{0,0,0}\rotatebox{90}{\makebox(0,0)[lt]{\lineheight{1.25}\smash{\begin{tabular}[t]{l}filter \end{tabular}}}}}%
    \put(0,0){\includegraphics[width=\unitlength,page=5]{pics/semrafiner_architecture.pdf}}%
    \put(0.19986512,0.12777237){\color[rgb]{0,0,0}\makebox(0,0)[lt]{\lineheight{1.25}\smash{\begin{tabular}[t]{l}RIT\end{tabular}}}}%
    \put(0,0){\includegraphics[width=\unitlength,page=6]{pics/semrafiner_architecture.pdf}}%
    \put(0.29011198,0.15260788){\color[rgb]{0,0,0}\makebox(0,0)[lt]{\lineheight{1.25}\smash{\begin{tabular}[t]{l}moving\end{tabular}}}}%
    \put(0.28802522,0.13322179){\color[rgb]{0,0,0}\makebox(0,0)[lt]{\lineheight{1.25}\smash{\begin{tabular}[t]{l}instance\end{tabular}}}}%
    \put(0.27226095,0.11102513){\color[rgb]{0,0,0}\makebox(0,0)[lt]{\lineheight{1.25}\smash{\begin{tabular}[t]{l}segmentation\end{tabular}}}}%
    \put(0,0){\includegraphics[width=\unitlength,page=7]{pics/semrafiner_architecture.pdf}}%
  \end{picture}%
\endgroup%

%% file: pics/radius_transformer_block.pdf_tex
\begingroup%
  \makeatletter%
  \providecommand\color[2][]{%
    \errmessage{(Inkscape) Color is used for the text in Inkscape, but the package 'color.sty' is not loaded}%
    \renewcommand\color[2][]{}%
  }%
  \providecommand\transparent[1]{%
    \errmessage{(Inkscape) Transparency is used (non-zero) for the text in Inkscape, but the package 'transparent.sty' is not loaded}%
    \renewcommand\transparent[1]{}%
  }%
  \providecommand\rotatebox[2]{#2}%
  \newcommand*\fsize{\dimexpr\f@size pt\relax}%
  \newcommand*\lineheight[1]{\fontsize{\fsize}{#1\fsize}\selectfont}%
  \ifx\svgwidth\undefined%
    \setlength{\unitlength}{245.71800232bp}%
    \ifx\svgscale\undefined%
      \relax%
    \else%
      \setlength{\unitlength}{\unitlength * \real{\svgscale}}%
    \fi%
  \else%
    \setlength{\unitlength}{\svgwidth}%
  \fi%
  \global\let\svgwidth\undefined%
  \global\let\svgscale\undefined%
  \makeatother%
  \begin{picture}(1,0.49243441)%
    \lineheight{1}%
    \setlength\tabcolsep{0pt}%
    \put(0,0){\includegraphics[width=\unitlength,page=1]{pics/radius_transformer_block.pdf}}%
    \put(0.51889151,0.44390591){\color[rgb]{0,0,0}\makebox(0,0)[lt]{\lineheight{1.25}\smash{\begin{tabular}[t]{l}$\mathbf{X}^{\text{in}}$\end{tabular}}}}%
    \put(0.52389264,0.03386312){\color[rgb]{0,0,0}\makebox(0,0)[lt]{\lineheight{1.25}\smash{\begin{tabular}[t]{l}$\mathbf{X}_{1}^{\text{b}}$\end{tabular}}}}%
    \put(0.35116063,0.24387954){\color[rgb]{0,0,0}\makebox(0,0)[lt]{\lineheight{1.25}\smash{\begin{tabular}[t]{l}radius transformer layer\end{tabular}}}}%
    \put(0.14962854,0.44390591){\color[rgb]{0,0,0}\makebox(0,0)[lt]{\lineheight{1.25}\smash{\begin{tabular}[t]{l}$\mathbf{P}^{\text{in}}$\end{tabular}}}}%
    \put(0.25300667,0.32187392){\color[rgb]{0,0,0}\makebox(0,0)[lt]{\lineheight{1.25}\smash{\begin{tabular}[t]{l}linear layer $\mathbf{W}^b_2 \in \RR^{D_1\times D_1}$ + GELU\end{tabular}}}}%
    \put(0,0){\includegraphics[width=\unitlength,page=2]{pics/radius_transformer_block.pdf}}%
    \put(0.25696232,0.38771659){\color[rgb]{0,0,0}\makebox(0,0)[lt]{\lineheight{1.25}\smash{\begin{tabular}[t]{l}linear layer $\mathbf{W}^b_1 \in \RR^{D\times D_1}$ + GELU\end{tabular}}}}%
    \put(0,0){\includegraphics[width=\unitlength,page=3]{pics/radius_transformer_block.pdf}}%
    \put(0.2478321,0.08667109){\color[rgb]{0,0,0}\makebox(0,0)[lt]{\lineheight{1.25}\smash{\begin{tabular}[t]{l}linear layer $\mathbf{W}^b_4  \in \RR^{D_1\times D_1}$ + GELU\end{tabular}}}}%
    \put(0,0){\includegraphics[width=\unitlength,page=4]{pics/radius_transformer_block.pdf}}%
    \put(0.2478321,0.15251377){\color[rgb]{0,0,0}\makebox(0,0)[lt]{\lineheight{1.25}\smash{\begin{tabular}[t]{l}linear layer $\mathbf{W}^b_3  \in \RR^{D_1\times D_1}$ + GELU\end{tabular}}}}%
    \put(0,0){\includegraphics[width=\unitlength,page=5]{pics/radius_transformer_block.pdf}}%
    \put(0.83168627,0.35236497){\color[rgb]{0,0,0}\makebox(0,0)[lt]{\lineheight{1.25}\smash{\begin{tabular}[t]{l}MLP\end{tabular}}}}%
    \put(0,0){\includegraphics[width=\unitlength,page=6]{pics/radius_transformer_block.pdf}}%
    \put(0.83168248,0.11422375){\color[rgb]{0,0,0}\makebox(0,0)[lt]{\lineheight{1.25}\smash{\begin{tabular}[t]{l}MLP\end{tabular}}}}%
  \end{picture}%
\endgroup%

%% file: pics/results_semrafiner.pdf_tex
\begingroup%
  \makeatletter%
  \providecommand\color[2][]{%
    \errmessage{(Inkscape) Color is used for the text in Inkscape, but the package 'color.sty' is not loaded}%
    \renewcommand\color[2][]{}%
  }%
  \providecommand\transparent[1]{%
    \errmessage{(Inkscape) Transparency is used (non-zero) for the text in Inkscape, but the package 'transparent.sty' is not loaded}%
    \renewcommand\transparent[1]{}%
  }%
  \providecommand\rotatebox[2]{#2}%
  \newcommand*\fsize{\dimexpr\f@size pt\relax}%
  \newcommand*\lineheight[1]{\fontsize{\fsize}{#1\fsize}\selectfont}%
  \ifx\svgwidth\undefined%
    \setlength{\unitlength}{245.71800232bp}%
    \ifx\svgscale\undefined%
      \relax%
    \else%
      \setlength{\unitlength}{\unitlength * \real{\svgscale}}%
    \fi%
  \else%
    \setlength{\unitlength}{\svgwidth}%
  \fi%
  \global\let\svgwidth\undefined%
  \global\let\svgscale\undefined%
  \makeatother%
  \begin{picture}(1,0.88719588)%
    \lineheight{1}%
    \setlength\tabcolsep{0pt}%
    \put(0,0){\includegraphics[width=\unitlength,page=1]{pics/results_semrafiner.pdf}}%
    \put(0.03493923,0.71875567){\color[rgb]{0,0,0}\rotatebox{90}{\makebox(0,0)[lt]{\lineheight{1.25}\smash{\begin{tabular}[t]{l}reference\end{tabular}}}}}%
    \put(0.04070594,0.52955514){\color[rgb]{0,0,0}\rotatebox{90}{\makebox(0,0)[lt]{\lineheight{1.25}\smash{\begin{tabular}[t]{l}RIT\end{tabular}}}}}%
    \put(0.04068607,0.1164318){\color[rgb]{0,0,0}\rotatebox{90}{\makebox(0,0)[lt]{\lineheight{1.25}\smash{\begin{tabular}[t]{l}GT\end{tabular}}}}}%
    \put(0.04068607,0.30053914){\color[rgb]{0,0,0}\rotatebox{90}{\makebox(0,0)[lt]{\lineheight{1.25}\smash{\begin{tabular}[t]{l}Ours\end{tabular}}}}}%
    \put(0,0){\includegraphics[width=\unitlength,page=2]{pics/results_semrafiner.pdf}}%
  \end{picture}%
\endgroup%